\documentclass{article}

\usepackage{corl_2018}
\usepackage{mathptmx}       % selects Times Roman as basic font
\usepackage{helvet}         % selects Helvetica as sans-serif font
\usepackage{courier}        % selects Courier as typewriter font
\usepackage{type1cm}        % activate if the above 3 fonts are
\usepackage{amsmath}
\usepackage{amssymb}
\usepackage{makeidx}         % allows index generation
\usepackage{graphicx}        % standard LaTeX graphics tool
\usepackage{multicol}        % used for the two-column index
\usepackage[bottom]{footmisc}% places footnotes at page bottom
\usepackage{enumerate}
\usepackage[all]{xy}
\usepackage{caption}
\usepackage{cite}
\usepackage[ruled]{algorithm2e}
\usepackage{bm}
\usepackage{mathrsfs}
\usepackage{graphicx}
\usepackage{epsfig}
\usepackage{pdfpages}
\usepackage{wrapfig}
\usepackage{lipsum}
\usepackage{times}
\usepackage{setspace}
\usepackage{cancel}
\usepackage{wrapfig}
\usepackage[rightcaption]{sidecap}
\usepackage{adjustbox}
\usepackage{xspace}
\usepackage{float}
\usepackage{hyperref}

\usepackage{subfigure}
\usepackage{lmodern}% http://ctan.org/pkg/lm
\newcommand{\latinword}[1]{\textsf{\itshape #1}}%

\newcommand{\ignore}[1]{}

\newcommand{\colour}[2]{\color{#1}{#2}\color{black} }
\newcommand{\mbfdot}[1]{\dot{\mbf{#1}}}

\newcommand{\norm}[1]{\left\Vert#1\right\Vert} % Norm
\newcommand{\abs}[1]{\left\vert#1\right\vert} % Absolute value
\newcommand{\bbm}{\begin{bmatrix}}
\newcommand{\ebm}{\end{bmatrix}}
\newcommand{\bma}[1]{\left[\begin{array}{#1}}
\newcommand{\ema}{\end{array}\right]}

\DeclareMathAlphabet{\mbf}{OT1}{ptm}{b}{n}
\newcommand{\mbs}[1]{{\boldsymbol{#1}}}
\newcommand{\mbfbar}[1]{{\bar{\mbf{#1}}}}

\newcommand{\beq}{\begin{equation}}
\newcommand{\eeq}{\end{equation}}
\newcommand{\bdis}{\begin{displaymath}}
\newcommand{\edis}{\end{displaymath}}
\newcommand{\beqarray}{\begin{eqnarray}}
\newcommand{\eeqarray}{\end{eqnarray}}
\newcommand{\beqarraynn}{\begin{eqnarray*}}
\newcommand{\eeqarraynn}{\end{eqnarray*}}

\newcommand{\p}{\partial}

\newcommand{\trans}{{\ensuremath{\mathsf{T}}}} % transpose
\newboolean{draft-mode}
\setboolean{draft-mode}{false}
\newcommand{\sidenote}[1]{\ifthenelse{\boolean{draft-mode}}{\marginpar{\tiny\raggedright\textsf{\hspace{0pt}#1}}}{}}
\DeclareRobustCommand{\arnote}[1]{\ifthenelse{\boolean{draft-mode}}{\textcolor{blue}{\textbf{AR: #1}}}{}}
\DeclareRobustCommand{\ernote}[1]{\ifthenelse{\boolean{draft-mode}}{\textcolor{cyan}{\textbf{ER: #1}}}{}}
\DeclareRobustCommand{\fhnote}[1]{\ifthenelse{\boolean{draft-mode}}{\textcolor{red}{\textbf{FH: #1}}}{}}

\makeatletter
\def\thanks#1{\protected@xdef\@thanks{\@thanks
        \protect\footnotetext{#1}}}
\makeatother

\title{A Data-Efficient Approach to Precise and Controlled Pushing}

\author{Maria Bauza$^{*}$, Francois R. Hogan$^{*}$ and Alberto Rodriguez% <-this % stops a space
\\ Department of Mechanical Engineering ---
    Massachusetts Institute of Technology \\ \tt\small $<$fhogan,bauza,albertor$>$@mit.edu\thanks{$^{*}$ Authors with equal contribution.}}

\begin{document}
\maketitle

%===============================================================================

\begin{abstract}
Decades of research in control theory have shown that simple controllers, when provided with timely feedback, can control complex systems. Pushing is an example of a complex mechanical system that is difficult to model accurately due to unknown system parameters such as coefficients of friction and pressure distributions. In this paper, we explore the data-complexity required for controlling, rather than modeling, such a system.

Results show that  a model-based control approach, where the dynamical model is learned from data, is capable of performing complex pushing trajectories with a minimal amount of training data (\textless 10  data points). The dynamics of pushing interactions are modeled using a Gaussian process (GP) and are leveraged within a model predictive control approach that linearizes the GP and imposes actuator and task constraints for a planar manipulation task. %Experimental results show that the Gaussian process captures well the frictional interactions between pusher and object and that a minimal amount of data (<10 datapoints) is required to achieve closed-loop performance of precise pushing tasks. 
\end{abstract}

% Two or three meaningful keywords should be added here

%===============================================================================

\section{Introduction}\label{sec:introduction}

Control theory has a rich history of designing reactive, reliable, and accurate controllers by leveraging simple and approximate dynamical models of the real world. Due to the nature of feedback, which reevaluates the state of the system in real-time, approximate dynamical models that  capture the essential properties of the system can be effective at applying corrective decisions. In contrast to traditional model-based control approaches, a growing trend in the robotics community is to learn control policies directly from experience, which typically rely on a very large quantity of training data to achieve good performance. This paper explores the data-complexity required to control manipulation tasks with a model-based approach, where the model is learned from data. We employ this methodology to the problem of pushing an object on a planar surface, and find that we can design effective control policies with  small data requirements (less than $10$ datapoints) while achieving  accurate closed-loop performance.

We are particularly interested in contact-rich robotic tasks where the dynamics are largely dominated by frictional interactions. Such tasks remain challenging for both model-based and learning-based control approaches. Classical physics-based control methods struggle to control such systems due to the non amenable nature of their motion equations which include hybridness and underactuation. 
%. In particular, modeling frictional interactions from first principles yields  complementary formulations due to the hybridness of contact, that are notoriously difficult to simulation, plan, and control \textcolor{blue}{[cite]}. 
While model-free approaches do not require a description of the motion model, they typically rely on large quantities of data that make their generalization to more complex tasks challenging. %Moreover, the common practice to discretize the action space often leads to policies that are jerky in practice, and do not fully use the precision and dexterity from today's robots.

In this paper, we investigate a planar manipulation problem where the goal is to control the pose of an object using a robotic pusher. Planar pushing is a minimal example of an underactuated manipulation task where the object motion is dominated by frictional interactions~\citep{Hogan2016}. The accurate modeling of planar pushing from first principles has proved difficult, due to unknown coefficients of friction and indeterminacies in the pressure distribution between the object and the surface where it slides. %Furthermore, previous research has shown that planar pushing exhibits stochastic behavior that deterministic models struggle to capture~\citep{bauza2017}. %We are interested in the degree of complexity of the model required for effective closed-loop control.    %In this task, the robot attempts to control the motion of the object by nonprehensile interactions. 

This paper aims to show that it is possible to  accurately control a mechanical system such as planar pushing, with a very small amount of experimental data, through a flexible control architecture that combines GP regression with MPC. While each of these pieces has been separately studied in the context of planar pushing, the novelty of the paper lies in: 1) the combination of both; and 2) the insight that a very small amount of data is sufficient to control the system (less than 10 data points). The contributions of the paper are:
\begin{enumerate}
    \item Model-based control policy where the pushing model is learned directly from data. We show that a model learned with Gaussian processes can  be effectively used in an MPC framework.
    \item Performance comparison between analytical and data-driven controllers. We combine both analytical and learned model of pushing within an MPC framework.
    \item Study of the data-complexity requirements needed to achieve stable control.
\end{enumerate}

A key result is that while around $200$ datapoints are sufficient to match the performance of the analytical controller, a much smaller number, in the order of $10$, already produces functional stable tracking behavior. A video showing our approach and the experiments can be found at \url{https://youtu.be/Z45O480pij0}.

%and $10$ datapoints can produce a stable tracking of the desired trajectories.

 %The key findings of this paper are
%  \begin{enumerate}
%      \item Learning a model for control is ``easy.''
%      \item Learning a model within a model-predictive control formulation has major data requirements advantages over model-free policy searches.
%      \item By formulating the model in velocity space, the dynamical hybridness inherent of frictional contacts is less of an issue.
%  \end{enumerate}

%5. \textcolor{blue}{Results show that a data driven approach ...}

%The main contribution of this paper is a learned model-based controller design approach to the pushing problem that relieves the controller design from explicit hybridness. 
This paper is structured as: planar pushing modeling, controller design, experimental results, and discussion. Both theory and results are presented for two different modeling strategies: analytical and data-driven. The analytical model refers to a pushing model derived from first principles using Newtonian and frictional mechanics. This approach leads to hybrid dynamics that makes feedback control design difficult \citep{Hogan2016}. The data-driven approach is based on a smooth Gaussian process model learned from data, which proves effective for control purposes.  %The planar pushing section outlines the theoretic framework employed in both approaches. used to  The controller design presents a model predictive controller that can be applied to both analytical and data-driven models. Finally, both controller designs are applied and compared on a experimental race track that is designed to test the accuracy and performance of both approaches. The 

\section{Related Work}

The problem of planar pushing has a rich literature  due to its theoretical and practical importance as one of the simplest nonprehensile manipulation problems. Since \citet{Mason1986} introduced the problem in 1986, there has been a wealth of research on its modeling, planning, and control. 

Early work on pushing focused on first principles models of planar pushing interactions. Due to  indeterminacies in the pressure distribution between an object and its support surface, \citet{Mason1986} introduced the voting theorem that can  resolve the direction of rotation of an object under an external pushing action without explicit knowledge of the pressure distribution. Following Mason's seminal work, several researchers have proposed practical models, most notably  \citet{Goyal1991}, who introduced the concept of limit surface and  \citet{lynch1992manipulation} that used it to model the dynamics of planar pushing. 

In recent years, researchers have turned to data-driven techniques to improve the accuracy of pushing interactions~\citep{Salganicoff1993,Walker2008,Lau2011,Meric2015,Zhou2016}. Of particular interest is  \citet{Zhou2016}, that  presents a physics-inspired data-driven model for systems with planar contacts. The algorithm approximates the limit surface as the level set of a convex polynomial.

%, which maps the frictional applied force to the object twist and is the state-of-the-art in data-efficient friction modeling in robotics.

The limit surface has proven to be  a valuable tool for simulation \citep{lynch1992manipulation}  and planning~\citep{Chavan_ISRR_2017}, but has remained challenging for controller design due to its hybrid nature, i.e. contacts can stick or slide. To address this issue, recent work has either restricted the control to predefined dynamical regimes~\citep{Zhou2017} or applied  heuristic methods to plan through different contact modes in real-time. Another approach has been to deploy data-driven methods to control pushing tasks~\citep{Arruda2017, Pinto2017, Zeng2018,Finn2016,Ebert2017,Agrawal2016}. %Such learning-based approaches have been successful to a degree but rely on large quantities of data. 

%Object to the goal using MPPI and learning dynamics of pushing~\citep{Arruda2017}.
%Learn how to push and grasp from RGB images~\citep{Zeng2018}
%A probabilistic planning framework for planar grasping under uncertainty~\citep{Zhou2017}.
%Learning  to  Push  by  Grasping:  Using  multiple  tasks  for  effective
%learning (Lerrel Pinto and Abhinav Gupta)
%Finn  et  al  used  an  auto-encoder  based  forward model  that  is  used  in  a  model  predictive  control  schema  to  find push actions based on image input. Unsupervised  learning  for physical interaction through video prediction.
%Learning to Poke by Poking: Experiential Learning of Intuitive Physics
%Self-Supervised Visual Planning with Temporal Skip Connections

\section{Planar Pushing Modeling }\label{sec:planar_pushing}

The dynamics of planar pushing are notoriously difficult to model due to uncertainties in the system's coefficients of friction and the indeterminacy in pressure distribution between the  object and the support surface. By assuming quasi-static interactions (negligible inertial forces) and an ellipsoidal limit surface,  \citet{lynch1992manipulation} derive an analytical mapping between pusher and object velocities. To circumvent the approximations made in \citet{lynch1992manipulation}, \citet{bauza2017} have recently employed a data-driven approach to capture the dynamics of the system without relying on uncertain parameters such as pressure distributions or coefficients of friction. This section details both the analytical and data-driven models used in subsequent sections for controller design.

\subsection{Analytical Model}
\label{sec:analytical_model}
\begin{figure*}[h]
\centering
\vspace{-5mm}
\subfigure[Planar pushing system with world frame \latinword{$\mathcal{F}_a$}  and body frame \latinword{$\mathcal{F}_b$}. We denote the length of the square as $a$.]
{
	\includegraphics[width=5.8cm]{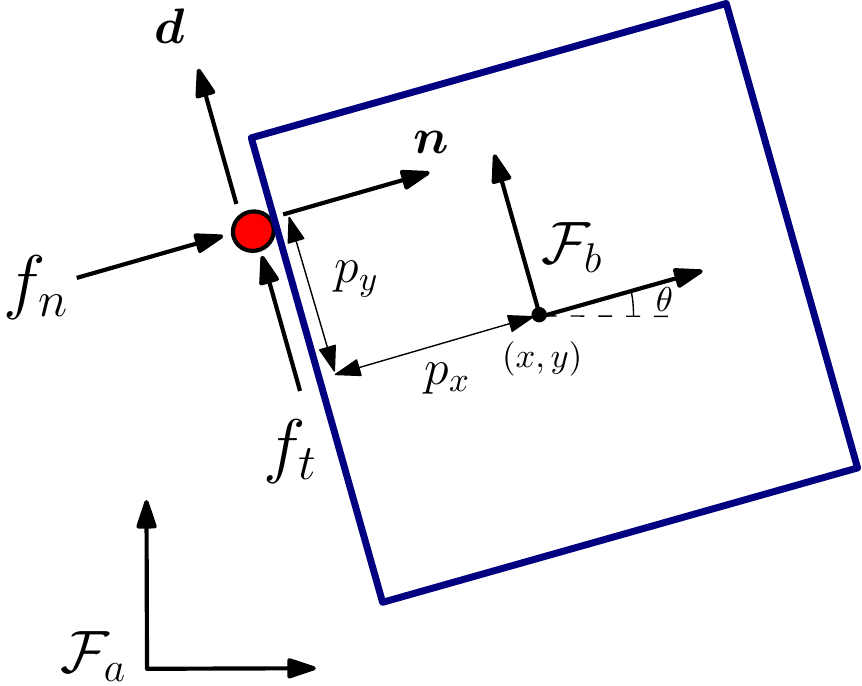} 
		\label{fig:pusher_FBD}
		\vspace{-10mm}
} 
\hspace{5mm}
\subfigure[Coulomb's frictional law. Three contact modes are defined with $\mu_p$ the coefficient of friction between the pusher and the object. 1. Sliding right: Friction acts as a force constraint, 2. Sticking: Friction acts as a kinematic constraint, 3. Sliding left: Friction acts as a force constraint. ]
{
			\includegraphics[width=6.5cm]{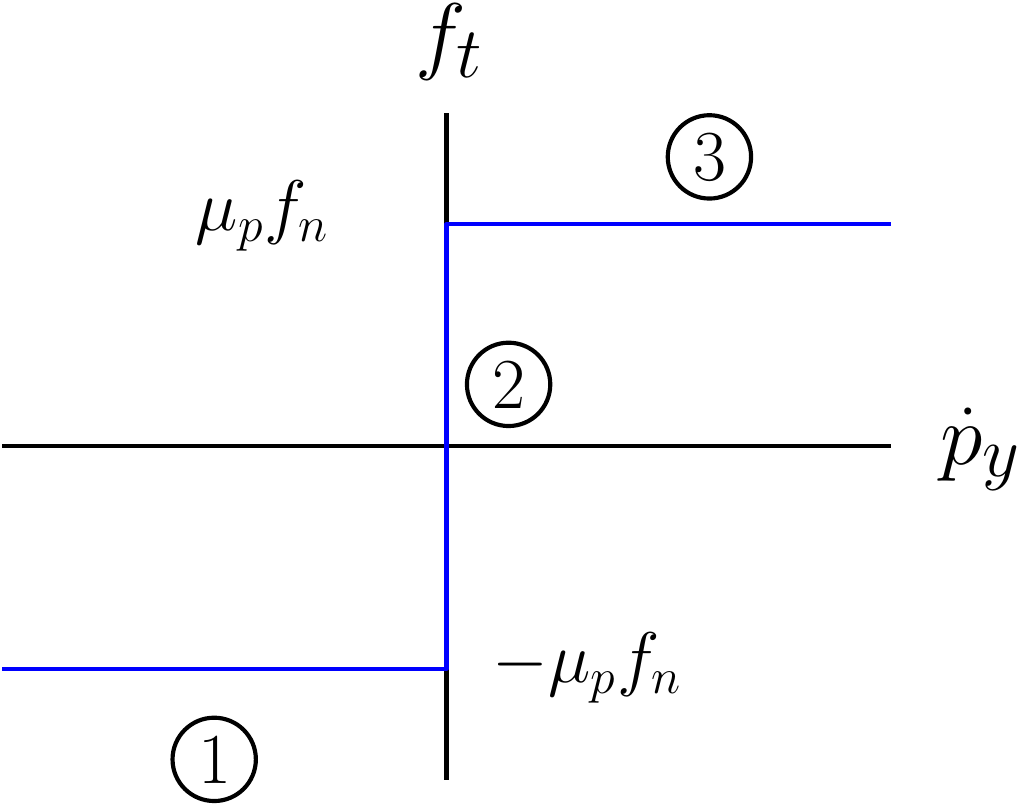} 
		 \label{fig:coulomb_model}
}
\caption{Problem description for the analytical pushing model. A key difficulty with the analytical model is that it exhibits different hybrid dynamics regimes due to the nature of Coulomb's friction. This complicates the feedback controller design.}
\vspace{-5mm}
\end{figure*}

Figure~\ref{fig:pusher_FBD} illustrates the planar pushing system, where $x,y, \theta$ denote the geometric center of the object and its orientation in the world frame.  The term $p_y$ relates the tangential distance between the pusher and the center-line of the object in the body frame.

 When the pusher interacts with the object, it impresses a normal force $f_n$, a tangential frictional force $f_t$, and torque $\tau$ about the center of mass. Assuming quasi-static interactions, the applied force causes the object to move in the perpendicular direction to the limit surface,  $H(\mbf{w})$,  as defined by \citet{Zhou2016}. As a result the object twist in the body frame  is given by $\mbf{t}=\mbs{\nabla}H(\mbf{w})$ where the applied wrench $\mbf{w} = [f_{n}\,\,f_{t}\,\,\tau]$ can be written as $\mbf{w} =  \mbf{J}^\trans\left(
\mbf{n}  f_{n}+ \mbf{d}{f}_{t}
\right)$
with $\mbf{n} = [1\,\,0]^\trans$, $\mbf{d} = [0\,\,1]^\trans$, and  $\mbf{J} = \bma{ccc}
1&0&-p_y\\
0&1&p_x
\ema$.
The system's motion equations are 
\beq
\mbfdot{x} = \mbf{f}_m(\mbf{x},\mbf{u}_m) =  \bma{cc}
\mbf{R}\,\mbf{t} \\ 
\dot{p}_y
\ema,
\hspace{3mm}
\mbf{R} = \bma{ccc}
\cos\theta&-\sin\theta&0\\
\sin\theta&\cos\theta&0\\
0&0&1
\ema,
\label{eq:motion_equations}
\eeq
where $\mbf{x} = [x\,\,y\,\,\theta\,\,p_y]^\trans
$ is the state vector
 and $\mbf{u}_m = [f_n\,\,f_t\,\,\dot{p}_y]^\trans$  the control input.  Due to the nature of physical interactions, the applied forces $f_n$, $f_t$ and the relative contact velocity $\dot{p}_y$ must obey frictional contact laws. Coulomb's frictional model is depicted in Fig.~\ref{fig:coulomb_model}, where three different regimes between the pusher and the object are identified: sliding right, sticking, and sliding left. The Coulomb's frictional model can be expressed using the mathematical constraints $\dot{p}_y=0$ and $\abs{f_t}<\mu_p\abs{f_n}$
 when the pusher is sticking relative to the object, $\dot{p}_y > 0$ and $f_{t} = \mu_p f_{n}$ when the pusher is sliding left relative to the object, and $\dot{p}_y < 0$ and $f_{t} = -\mu_p f_{n}$ when the pusher is sliding right relative to the object. As investigated in \citet{Hogan2016}, these discontinuous constraints in the dynamics lead to a hybrid system that makes controller design challenging.

\subsection{Data Driven  Model}

As an alternative to the analytical model, we consider the data-driven approach proposed by  \citet{bauza2017} that better captures the complex frictional interactions between pusher,  object, and support surface. \citet{bauza2017} showed that as few as $100$ samples are enough to train a Gaussian process (GP) to surpass the accuracy of the analytical model.

We train a GP to model for each output using a zero mean prior and the Automatic Relevance Determination (ARD) squared exponential kernel function $
{k}(\mbf{x}, \mbf{x}') ={ \sigma}_f^2 \text{exp}(-(\mbf{x}-\mbf{x}')^T \mbs{\Lambda}^{-1} (\mbf{x}-\mbf{x}')) 
$
where $\sigma_f^2$ is the signal variance and $\mbs{\Lambda}$ is a diagonal matrix with the estimated characteristics lengths of each input dimension \citep{Rasmussen2006}. 

\iffalse
As a result, the predicted value of $f$ for a new input $\mbf{x}^*$ is

$$
f(\mbf{x}^*) = \mbf{k}_*(\mbf{K} + \mbf{R})^{-1}\mbf{y}
$$

where $\mbf{k}_*$ represents the vector $[\mbf{k}_*]_i = {k}(\mbf{x}_i, \mbf{x}^*)$, $\mbf{K}$ is a matrix such that $[\mbf{K}]_{i,j} = {k}(\mbf{x}_i, \mbf{x}_j)$, $\mbf{R}$ is a diagonal matrix with the noise of the system at each input $\mbf{x}_i$, and $\mbf{y}$ is the vector of noisy observations $[\mbf{y}]_i = \mbf{y}_i$. When learning a GP, the hyperparameters $\sigma^2_f$, $\mbs{\Lambda}$ and $\mbf{R}$ have to be optimized while the other variables can be directly computed from the training data. 
\fi
To collect data, the robot executes pushes with random initial contact position $p_y$ and direction $\beta$ as in \citet{Yu2016}. The angle $\beta$ describes the  orientation of the pusher relative to the object body frame where $\tan{\beta} = \frac{v_n}{v_t}$, and  $v_n$, $v_t$ denote the pusher velocity in the body frame. The learning problem is defined as:

\textbf{Inputs}: $[p_y \,\, \beta]^\trans$, as defined above.

\begin{figure}[h]
\centering
{
		\includegraphics[ width=15cm]{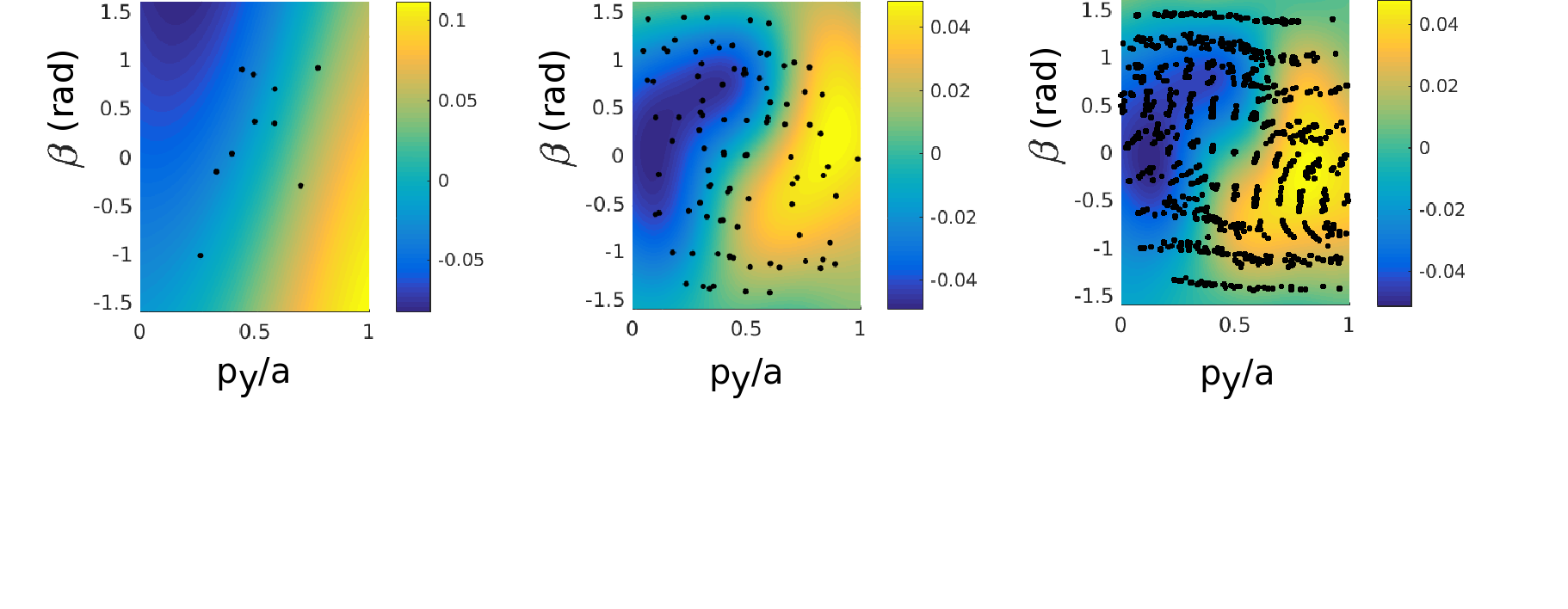} %
}
\centering
\vspace{-22mm}
\caption{Training data (dots) and learned model for the object's change in orientation, $\Delta\theta_b$ (rad). From left to right the number of datapoints is 10, 100 and 1000. We observe that the model complexity and accuracy increases with the number of training data. } \label{fig:3D_model}

\end{figure}

\textbf{Outputs}: $\Delta \mbf{x}_b = [\Delta x_b  \,\, \Delta y_b \,\, \Delta  \theta_b]^\trans$, where $\Delta x_b$, $\Delta y_b$, and $\Delta \theta_b$ represent the displacement of the object's center and change in orientation in the body frame for the duration of the push $\Delta t$. Figure~\ref{fig:3D_model} shows the model obtained for $\Delta \theta_b$ depending on the training datapoints.

By leveraging the quasi-static assumption, which neglects inertial effects, the model is learned for a predetermined velocity $V_{nom}$ and scaled proportionally with the velocity of the pusher to recover the object velocity in the body frame  as  $
\mbfdot{x}_b = \frac{\norm{\mbf{v}_p}}{V_{nom} \Delta t} \Delta \mbf{x}_b$, where $\mbf{v}_p=[v_n\,\,v_t]^\trans$ is the pusher velocity in the body frame.

%---------------------------

We write the data-driven motion equations in a similar form to~\eqref{eq:motion_equations}. The velocity of the pusher relative to the object is resolved in the body frame as 
\beq
\bma{cc}
\dot{p}_x\\ \dot{p}_y
\ema =\mbf{v}_p- \mbf{J}\mbfdot{x}_b.
\label{eq:kinematics}
\eeq
 The data-driven motion equations are
\beq
\mbfdot{x} = \mbf{f}_d(\mbf{x}, \mbf{u}_d)=\bma{cc}
\mbf{R} \mbfdot{x}_b\\ \dot{p}_y
\ema,
\label{eq:data_driven_motion_equations}
\eeq
where the control input for the data-driven model is $\mbf{u}_d = \mbf{v}_p$. 

\section{Controller Design}\label{sec:controller_design}

This section presents the  feedback policy design used for  real-time control of planar pushing. We control both the analytical and the learned models with a model predictive control (MPC) framework due to its flexibility to the algebraic form of the model, and the possibility to enforce state and action constrains. The model predictive controller acts by simulating the model forward and finding an open loop sequence of control inputs that brings the system close to a desired trajectory. By resolving this  optimization in real-time and applying the first action determined in the control sequence, this strategy can act as an effective closed-loop stabilizing policy. 

\noindent \textbf{Model Predictive Control (MPC)}: Given the current error state $\mbfbar{x}_0$ and a nominal trajectory ($\mbf{x}_i^\star$, $\mbf{u}_i^\star$), solve 
\vspace{-2mm}
\beq
% \hspace{-5mm}
\begin{aligned}
& \underset{\mbfbar{x}_i,\hspace{1mm} \mbfbar{u}_i}{\text{min}}
& &  \mbfbar{x}_N^\trans\mbf{Q}_N\mbfbar{x}_N + \sum_{i=0}^{N-1} \left(\mbfbar{x}_{i+1}^\trans\mbf{Q}\mbfbar{x}_{i+1} + \mbfbar{u}_i^\trans \mbf{R}\mbfbar{u}_i
\right) \\
& \text{subject to}
& & {\mbfbar{x}}_{i+1} ={\mbfbar{x}}_{i}+ h\left[\mbf{A}_i{\mbfbar{x}}_{i}+ \mbf{B}_i{\mbfbar{u}}_{i}\right]\, \hspace{5mm}\text{(Linearized Motion Equations)}, 
\\
&& &\mbf{x}_i\in \, \text{\latinword{$\mathcal{X}$} \,\,(State Constraints)}, \\
&& &\mbf{u}_i\in \text{\latinword{$\mathcal{U}$} \,\,(Input Constraints)}, 
\end{aligned}
\label{mpc_miqp}
\eeq
\vspace{-4mm}

\noindent with integration time step $h$, $\mbfbar{x}_i = \mbf{x}_i - \mbf{x}_i^{\star}$ and $\mbfbar{u}_i = \mbf{u}_i - \mbf{u}_i^{\star}$.   The terms $\mbf{Q}$, $\mbf{Q}_N$, and  $\mbf{R}$ denote weight matrices associated with the error state, the final error state, and the control input. The optimization is performed by linearizing the dynamics of the system about a desired nominal trajectory with $\mbf{A}_i  = \frac{\p \mbfdot{x}}{\p \mbf{x}}|_{\mbf{x}_i^\star, \mbf{u}_i^{\star}}$ and $\mbf{B}_i  = \frac{\p \mbfdot{x}}{\p \mbf{u}}|_{\mbf{x}_i^\star, \mbf{u}_i^{\star}}$. The nominal trajectory is computed using the analytical model with sticking interactions to avoid hybridness as done in \citet{zhou2017pushing}.
%For the pusher-slider system, the nominal trajectory is computed using nonlinear trajectory optimization \textcolor{red}{[cite]} using the analytical model. By limiting the nominal trajectory to sticking interactions, the optimizing does not suffer from system hybridness and can be found using standard gradient descent algorithms such as matlab 's $\texttt{fmincon}$ function. For more details on how to design the nominal trajectory, see \cite{Hogan2018}. 

The model predictive approach offers the flexibility to test the same controller design on both the analytical and the data-driven model. The controller design formulation for both settings is described below.

%\colour{red}{TODO:} add somewhere the definition of the time step $h$ and maybe also of A and B matrices.

\subsection{Analytical Model}
The analytical model uses the control input $\mbf{u}_m = [f_n\,\,f_t\,\,\dot{p}_y]$ along with the motion equations~\eqref{eq:motion_equations}. %, which are linearized at the desired trajectory and controls: $\mbf{A}_i  = \frac{\p \mbf{f}_m}{\p \mbf{x}}|_{\mbf{x}_i^\star, \mbf{u}_i^{\star}}$ and $\mbf{B}_i  = \frac{\p \mbf{f}_m}{\p \mbf{u}}|_{\mbf{x}_i^\star, \mbf{u}_i^{\star}}$. 
We include a constraint on $\abs{p_y}$ to keep the pusher within the object's edge.

Due to the nature of frictional contacts, as described earlier in Section~\ref{sec:analytical_model}, the input constraints are hybrid and not amenable to conventional MPC designed for continuous systems. In this paper, we follow the Family of Modes (FOM) heuristic, as introduced in \citet{Hogan2016}, to address this problem. %The FOM approach consists in optimizing over a finite number of fixed contact mode sequences. The input constraints associated with each determined contact mode are enforced during the MPC time horizon.

\subsection{Data Driven  Model}

The data-driven model is more amenable than the analytical model for MPC as it presents continuous differential equations. As such, no particular care needs to be taken with regard to system hybridness and selecting mode sequences. The control input is given directly by the velocity of the pusher in the body frame of the object $\mbf{u}_d = \mbf{v}_p$ along with the motion equations~\eqref{eq:data_driven_motion_equations} linearized about the nominal trajectory. 

To address the stochastic nature of GPs, we make use of the certainty equivalent approximation \citep{bertsekas1995}, which acts by settings random values with their expected value during the optimization process. In the case of the pusher-slider system modeled with GPs, this implies that the mean of the dynamics are propagated forward by setting the state noise value to $0$.  This approximation is computationally beneficial since it converts a stochastic optimization problem into a deterministic one. This approximation has been shown to produce good results for linear systems, where the certainty principle is optimal for systems with additive Gaussian noise.
	
\section{Results}\label{sec:results}

This section evaluates the performance of the analytical and data-driven controllers based on their ability to follow a given track. The purpose of the task is to accurately control the motion the object using a point robotic pusher about a desired timed trajectory.

\begin{figure*}[h]
\centering
\subfigure[8-track. Two identical circles of radius $150$mm that intersect. This trajectory is tested at target velocities of $20$mm/s and $80$mm/s.]
{
			\includegraphics[width=6.3cm]{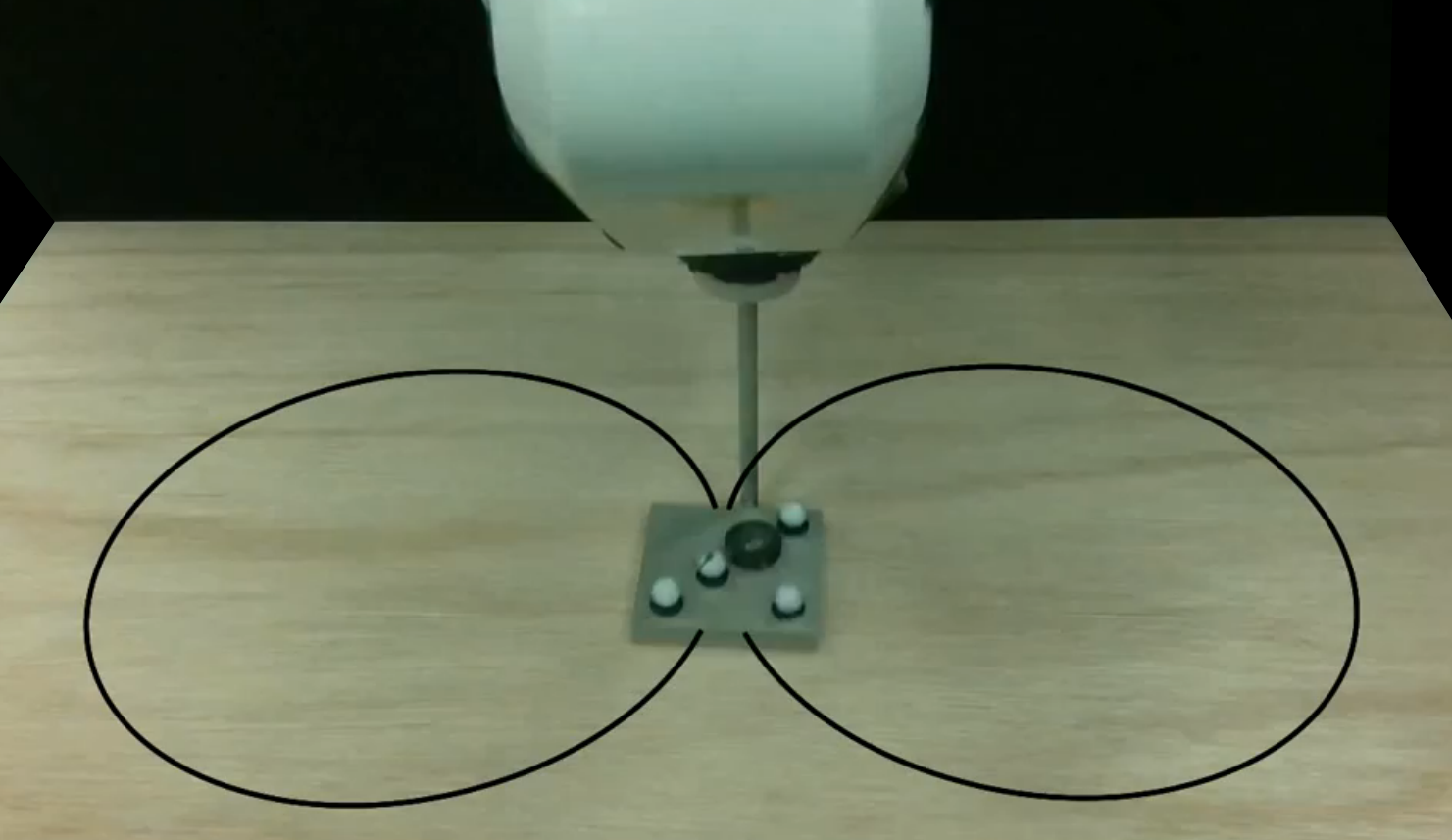}
		\label{fig:8_track}
}
\hspace{0mm}
\subfigure[Square track. Four straight lines where the borders between adjacent lines are connected with quarter circles of $80$mm radius. This trajectory is performed at a test velocity of $50$mm/s.]
{
			\includegraphics[width=6.9cm]{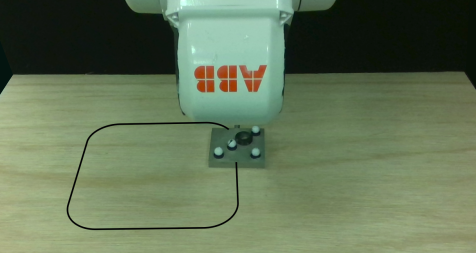} 
		 \label{fig:square_track}
}
\caption{\footnotesize{Experimental setup. A metallic rod  is attached to an ABB IRB 120 industrial robotic arm to push an object. The pose of the slider is tracked using a Vicon camera system.}} 
\label{fig:target_trajectories}
\vspace{-5mm}
\end{figure*}

\subsection{Experimental Setup}

Figure~\ref{fig:target_trajectories} depicts the robotic setup and the two target trajectories investigated during the experiments. We use an industrial robotic manipulator (ABB IRB 120) along with a Vicon camera system to track the position of the object as in \citep{Hogan2018}. The material used for the support surface is plywood with an estimated frictional coefficient of $\mu_g  = 0.35$. The estimated coefficient of friction of the  pusher-object interactions is $\mu_p = 0.3$ and the object  is a square of length $90$mm and $0.827$ kg of mass.

The parameters of each controller are tuned to obtain their best performance on the 8-track trajectory at $80$ mm/s. For the data-driven controller, the parameters are only optimized for the model with 5000 data points. The MPC parameters for the analytical model are $\mbf{Q} = [6000, 3000, 10, 0]$, $\mbf{Q}_N = \mbf{Q}$,  associated with the state $\mbf{x} = [  x\,\,  y\,\, \theta \,\, p_y]^\trans$ and  $\mbf{R} = [0.1, 0.001, 0.001]$ associated with  $\mbf{u}_m = [f_n\,\,f_t\,\,\dot{p}_y]^\trans$. For the learned model, we use $\mbf{Q} = [6000, 3000, 10, 3000]$, $\mbf{Q}_N = \mbf{Q}$,  and $\mbf{R} = [10, 0.001]$, associated with $\mbf{x} = [x\,\,y\,\,\theta\,\,p_y]^\trans$ and $\mbf{u}_d = \mbf{v}_p$, respectively. Both controllers consider time increments of $h = 0.01$s and $N = 35$ time steps.

\subsection{Model comparison}
\begin{figure*}[t]
\centering
\subfigure[8-track.]
{
			\includegraphics[width=6.5cm]{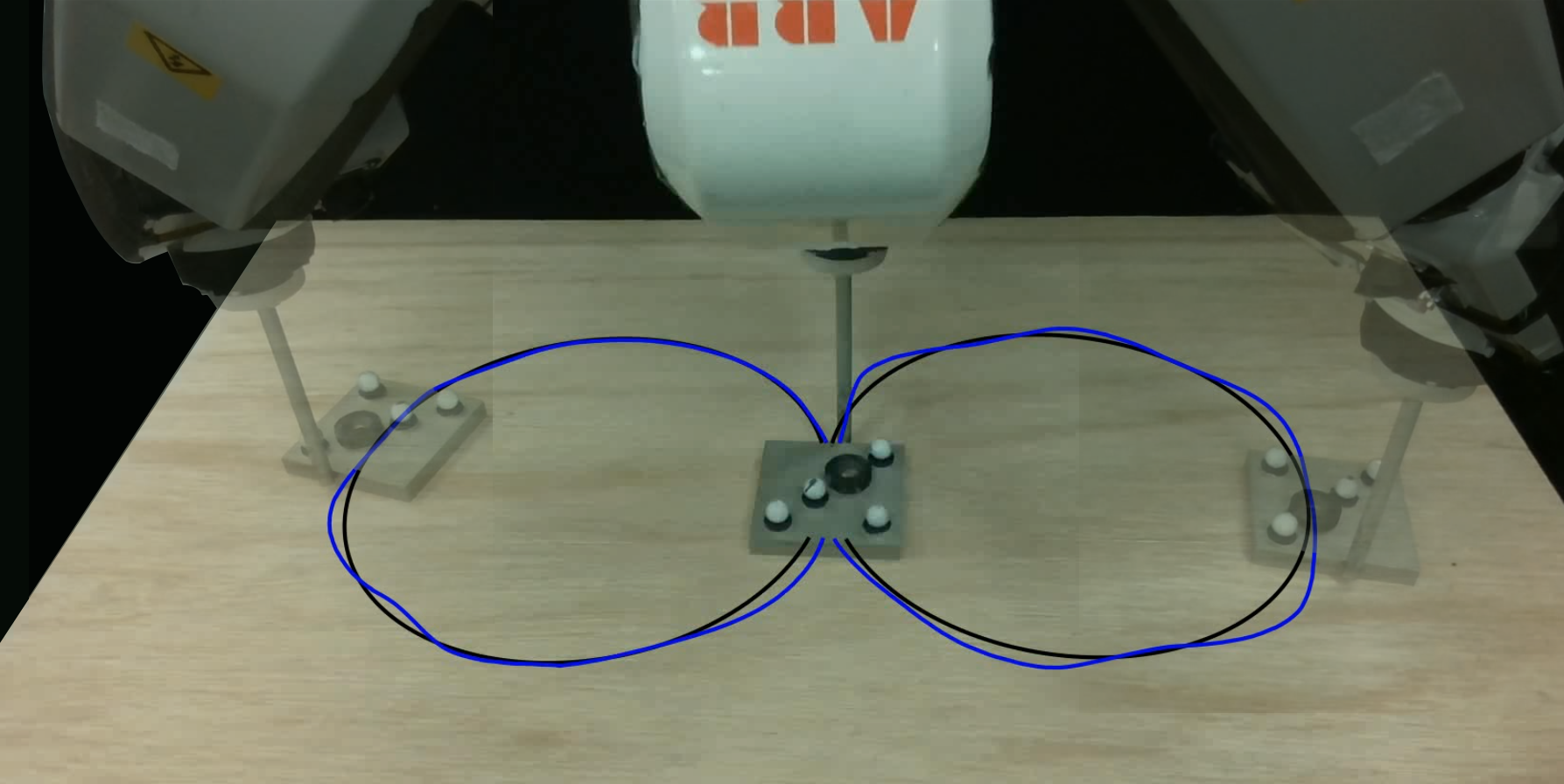}
		\label{fig:8_track_comparison}
}
\hspace{0mm}
\subfigure[Square track. ]
{
			\includegraphics[width=6.2cm]{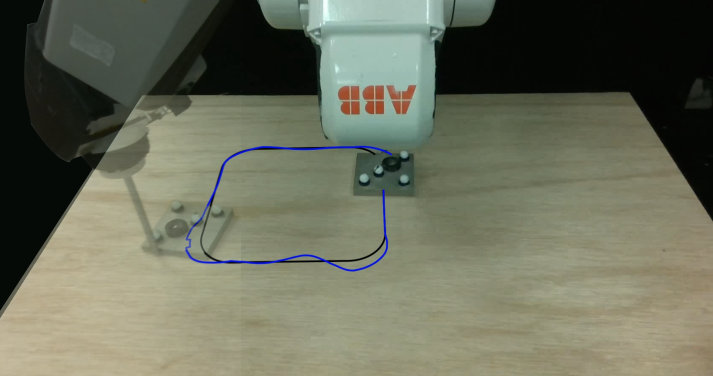} 
		 \label{fig:square_track_comparison}
}
\caption{Desired trajectory (black) compared with the motion followed by the object's geometric center (blue) using the analytical controller at 80 and 50 mm/s respectively. The average error for the 8-track example is 9.00mm and 6.05mm for the square.} \label{fig:traj_comparison} 
\vspace{-5mm}
\end{figure*}

Table~\ref{table:results} compares the performance of the analytical and the data-driven controller designs on a series of trajectory tracking problems. The performance of each controller is measured by computing the average mean squared error over the duration of the experiment between the desired trajectory and the actual motion of the object's geometric center (see Fig.\ref{fig:traj_comparison}). In this section, we conduct four benchmark experiments for each controller:
1. low tracking velocity ($20$ mm/s) without external perturbations, 2. high tracking velocity ($80$mm/s) without external perturbations, 3. high tracking velocity ($80$mm/s) with external perturbations, and 4. square tracking at $50$mm/s. Two perturbations types are considered in order to test the robustness of the controllers: tangential and normal. Tangential perturbations are applied laterally to the motion of the object by perturbing the initial position of the contact point from its desired position. Normal disturbances are applied orthogonal to the object's motion  by detaching the object away from the robotic pusher by $30$mm.

%---------------------------
 
%%%%%%%%%%%%%%% begin table   %%%%%%%%%%%%%%%%%%%%%%%%%%
\begin{table}[h]
\caption{Controller performance comparison}
\vspace{-5mm}
\begin{center}
\label{table:results}
\begin{tabular}{l l l l l}
& &  \\ 
Trajectory   & Error (Analytical)  & Error (Data-Driven)\\
\hline
8-track no perturbation, v$=80$mm/s & $ 9.56$ mm & $ 8.50$ mm \\
8-track no perturbation, v$=20$mm/s & $ 2.89$ mm & $ 6.53$ mm \\
8-track normal perturbation, v$=80$mm/s & $ 11.10$ mm & $ 8.52$ mm \\
8-track tangential perturbation, v$=80$m/s & $ 12.37$ mm & $ 9.28$ mm \\
Square trajectory, v$=50$mm/s & $ 4.95$ mm & $ 6.60$ mm \\
% \colour{red}{integration step size} & h & $0.01$
\end{tabular}
\end{center}
\vspace{-5mm}
\end{table}

Table~\ref{table:results} summarizes the tracking performance results for the analytical and data-driven (5000 datapoints) controllers. Both controller designs successfully achieve closed-loop tracking within $10$mm accuracy when no external perturbations are applied. It is worth noting that although the data-driven model only performs marginally better than its model-based counterpart, its controller design is much simpler to implement as it relies on a continuous dynamical model and doesn't suffer from discontinuous dynamics.

% We can observe that:

% \begin{itemize}
%     \item Parameters can be tuned to get good performances for both controllers
%     \item Performance is similar in most cases
%     \item GP seems better at recovering from perturbations (recall parameters are fixed). When the perturbation is to detach the object from the pusher, GP corrects almost immediately. 
%     \item GP seems to do better at lower velocities, but GP tuned properly should do equally good. (hypothesis)
% \end{itemize}

\subsection{Influence of data}

\begin{figure}[h]
% \vspace{-8mm}
\centering
  \includegraphics[width=12.cm]{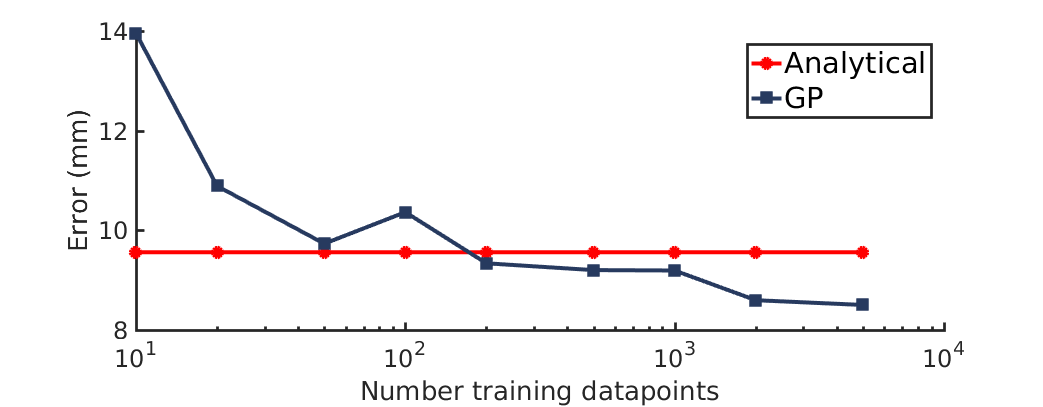}
\caption{\footnotesize{Average error on the 8-track at $80$mm/s depending on the model and the number of training points.}} 
\label{fig:error_vs_data}
%\vspace{-5mm}
\end{figure}

To evaluate the amount of data required to perform the trajectory tracking task, we run the controller for the data-driven model with an increasing quantity of  training data: $N=\{ 10, 20, 50, 100, 200, 500, 1000,2000, 5000\}$. We consider the case of high velocity ($80$mm/s) as it is the most challenging for the controller design. Figure~\ref{fig:error_vs_data} shows that around two hundred points are sufficient to get an accuracy equivalent to that of the analytical model. 
%and  high accuracy and that performance  plateaus after adding sufficient data.

When decreasing the amount of data, the control parameters are tuned for the model with $5000$ data points and kept equal across models. The data for the reduced datasets takes subsets such that smaller datasets are always contained within bigger ones. 

%Fixed the parameters of the controller, we obtained that after decreasing the total number of datapoints used to learn the control we can still track the desired trajectory and the controller is stable. This subset of 10 points is chosen completely at random. For smaller random datasets the controller became unstable. However, as discussed in Section~\ref{sec:discussion}, when we constrained the set of training points to be in a determined region the controller work with only 5 training points and without having to tune the controller parameters.

Figure~\ref{fig:error_vs_data} reports the effect of increasing the size of the data to train the GP model. Results show that the tracking performance increases as the model accuracy improves. Perhaps most surprising is that closed-loop performance was possible for the data-driven model with as few as $10$ data points with an accuracy of $14$ mm. This illustrates that feedback control can work with very simple dynamical models that capture the essential behavior of the dynamics. By continuously reassessing the state of the object  and recomputing the control action in real-time, the controller can act as to correct previous mistakes and track the desired trajectory. 

For GP models trained on random datasets with less than $10$ points, the controller was unstable, i.e., unable to complete the entire lap while remained in contact with the pusher. The data generated for training the GP model in Fig.~\ref{fig:error_vs_data} is at random within the available dataset of experimental pushes.  However, interestingly,  as later discussed in Section~\ref{sec:discussion}, when we carefully select the data used for training the GP, the controller can work with as few as $5$ training points. This result on the lower limit for closed-loop stable tracking was unexpected and shows that data-efficiency can be achieved through a model-based control approach, where the model is directly learned from data. 

%%%%%
%  change data: high velocity , perturbations with and without, 
%%%%%
%  
%%%%
\section{Discussion}\label{sec:discussion}

This paper explores a data-driven approach to control planar pushing tasks. Results show that learning the dynamics of the system from data and controlling the resulting model can be a data-efficient approach, and can achieve reactive and accurate robot/object interactions. We choose an algebraic form for the model that is continuously differentiable (GP), which is more amenable to apply tools from control theory. The hyperparametrization of the GP model as a function of the data ensures that even if the task has hybrid dynamics, the hybridness is only explicit in the data distribution but not in the algebraic form. In practice, we have found that a minimal amount of data can lead to a stable controller, implying that approximate models of contact interactions can be effective when combined with feedback. Given a set of fixed control parameters, the presented  data-driven controller design was able to track both the 8-track and square trajectories show in Fig.~\ref{fig:target_trajectories} at varying speeds, and for different amounts of data and perturbations for the case of the 8-track. This ability to generalize to new tasks is an advantage of model-based control formulations where the model learned can be adapted to new scenarios by leveraging physics.

% This works presents ...
% \begin{itemize}
%     \item A data-driven approach for controlling planar pushing motions.
%     \item Results show that this approach can achieve similar or improve the accuracy of physics-based controllers while being data efficient and without considering explicitly the hybridness or the underactuation of the problem.
%     \item In practice, minimal amounts amounts of data have been found to produce stable controllers. This might imply that accurate modeling is not paramount in the presence of control.
%     \item Given the same control parameters, our approach has been capable of handle several perturbations and work for different desired trajectories. This suggest that our control solution is robust as it can handle modifications in the system.
% \end{itemize}

In this paper, we aimed to answer questions regarding the learning and control representations that are more appropriate for reactive manipulation tasks. What is the level of complexity that should be captured by the motion model? How much data is required? Should system hybridness be explicitly included in the motion model? Our key findings are:
 \begin{enumerate}
     \item Learning a model for control is easier than learning a model for accurate simulation. We are able to get high performance control using a model trained on as little as $5$ hand-selected datapoints or $10$ datapoints selected at random. This result indicates what it is not surprise to the control community, that approximate simple models can enable powerful control mechanisms. We have shown that this simple models can be learned from data.
     \item The combination of GP model learning and model predictive control yields a practical and data-efficient framework to learn and control motion models. By stabilizing the motion of an object about a predetermined trajectory, MPC only requires the learned model to be accurate to first order by exploiting feedback. 
     \item By formulating the model in velocity space, the hybridness inherent of frictional contacts (see Fig.\ref{fig:coulomb_model}) softens  as the implicit dependence on reaction forces is hidden. This smoothing of the dynamics offers major controller design advantages without compromising the accuracy of the resulting controllers.
 \end{enumerate}
 
%  \colour{red}{NOTE:} the next paragraph seems a bit disconnected or maybe it should go up in the discussion.
% By leveraging known properties of physical interactions, such as quasistaticness, we are able to learn a data-efficient pushing model. Most importantly, by formulating the learning problem in velocity space, we are able to learn a smooth mapping between robot pushing velocities and object velocities that is no longer hybrid as the implicit dependence on reaction forces is hidden. We show that this smoothing of the dynamics offers major controller design advantages without compromising the accuracy of the resulting controllers.

% What we have observe/ questions we can answer...
% \begin{itemize}
%     \item Do we really need to us a GP to learn the dynamics? It seems that easier approaches and learning tools would be sufficient as only 10 data points are enough.
%     \item We have selected only 5 points from a narrow set of regions (forming in the input space the shape of a 5 in the dice) and we have made the controller converge 4 out of 5 times with high accuracy, around $9$mm error, as can be seen in Fig.~\ref{fig:traj_comparison}. This suggests that expert knowledge of the dynamics maybe sufficient to get a controller that works. Note: the 5 points selected cover different regions of the input space that correspond to negative and positive displacements and orientation changes. 
% \end{itemize}

A limitation of the proposed methodology is that it relies on tracking a nominal trajectory both in state and actuation spaces.  These are required when linearizing the system's dynamics and are obtained in this paper by relying on planning with respect to the analytical model. %In this work, we obtain the nominal trajectory and the sequence of optimal actions by assuming that the dynamics are given by the analytical model, as performed in \citep{Hogan2018}. In most problems however this offline knowledge is not readily available and the controller design proposed in this work would need to be augmented with a trajectory planner that finds feasible trajectories using the learned dynamics. 
Another limitation is that the contact state between the object and the robot is always assumed to be in a contact phase. As such, the controller cannot reason about separation and does not have the ability to switch sides to have better controllability on the object. These generalization would yield a combinatorial input space which would be difficult to handle by a regular GP.

To further improve the performance of the system, we believe that online adaptation of the nominal trajectory (feedforward control) by performing iterative learning control \citep{bristow2006survey} can achieve more aggressive and accurate pushing actions. Most importantly, we are interested in applying the learning control methodology presented in this paper to manipulation problems of higher complexity, with more contact formations and higher degrees of freedom. 

%Generalization would add complexity to the GP model, which benefits from a continuous input/actuation space.

% \begin{itemize}
    % \item How do we get even higher accuracy? How do we make a controller that can follow almost perfectly the desired trajectory?  
    % \item We know our approach does the following approximations: linearizes the dynamics, requires a nominal trajectory which we built using the analytical model, assumes that the pusher is always in contact with the object.
%     \item Options to improve the system: online adaptation of the nominal trajectory, changing the control strategy to avoid linearizing the dynamics or better regulate when the pusher is in contact with the object. Reinforcement learning approaches could also help to solve the problem by directly learning a policy. Internal note: this is an important topic of the discussion because it opens the door to new contributions where we refine the controller approach of the nominal trajectory online to improve the system performance
% \end{itemize}

% We do not talk about the influence of the surface (plywood) or the vicon cameras.
%===============================================================================

% The maximum paper length is 8 pages excluding references and acknowledgements, and 10 pages including references and acknowledgements

\clearpage
% The acknowledgments are automatically included only in the final version of the paper.
%\acknowledgments{If a paper is accepted, the final camera-ready version will (and probably should) include acknowledgments. All acknowledgments go at the end of the paper, including thanks to reviewers who gave useful comments, to colleagues who contributed to the ideas, and to funding agencies and corporate sponsors that provided financial support.}

%===============================================================================

% no \bibliographystyle is required, since the corl style is automatically used.
\bibliography{example}  

\end{document}